\documentclass[a4paper,12pt]{article}

\usepackage[english]{babel}

\usepackage[a4paper,top=3cm,bottom=2cm,left=3cm,right=3cm,marginparwidth=1.75cm]{geometry}

\usepackage{reduced_arxiv}
\usepackage{amsmath}
\usepackage{amssymb}
\usepackage{graphicx}
\usepackage[colorinlistoftodos]{todonotes}
\usepackage[colorlinks=true, allcolors=blue]{hyperref}
\usepackage{pdfpages}
\usepackage{color}
\usepackage[nottoc,numbib]{tocbibind}

\newcommand{\LL}{\mathcal{L}}
\newcommand{\NN}{\mathcal{N}}
\newcommand{\DD}{\mathcal{D}}
\newcommand{\EE}{\mathbb{E}}
\newcommand{\RR}{\mathbb{R}}

\DeclareMathOperator{\rk}{rk}
\DeclareMathOperator{\st}{s.t.}
\DeclareMathOperator*{\Argmin}{Arg\,min}

\title{An essay on optimization mystery of deep learning}
\author{
    {Eugene A. Golikov}
    \\
    Neural Networks and Deep Learning Lab., MIPT
    \\
    \texttt{golikov.ea@mipt.ru}
    \\
    \texttt{golikov.e.a000@gmail.com}
}
\date{}

\begin{document}

\maketitle

\section{Introduction}

Despite its huge empirical success, deep learning still preserves many features of alchemy~\cite{Rahimi_2017}: progress in this field is obtained mainly by trial and error, and our intuition about how do neural networks actually work often misleads us.

Alchemy, in order to become usual chemistry, needs a theoretical ground. For now, a solid theoretical ground for deep learning is lacking, however, fortunately, many pieces of theory appeared from different directions during several past years. The purpose of this essay is not to provide a comprehensive review, but to draw connections between some works on this topic. The list of works mentioned here is by no means representative, or, all the more so, complete.

Since the theory of deep learning is lacking, some features of neural networks learning seem "mysterious". We emphasize two mysteries of deep learning:
\begin{enumerate}
    \item \textbf{Generalization mystery.} It is very common for contemporary neural networks to have many more parameters than the number of training examples at hand. Having huge abundance of parameters results in existence of "bad" local minima in terms of test error \cite{Zhang_et_al_2016}. Classic generalization bounds based on VC-dimension are pessimistic, and consequently vacuous in this case. However, stochastic gradient descent seems to avoid these bad local minima. Hence new, neural-net-specific, bounds have to be developed.
    
    \item \textbf{Optimization mystery.} It is indeed surprising, why (stochastic) gradient descent does not get trapped into "bad" local minima in terms of test error. However, it is also quite surprising, why gradient descent (a local optimization method!) does not get trapped into bad local minima in terms of \emph{train} error. It seems true that, since the number of parameters in modern neural networks is usually much greater than the number of training samples, there exists such parameter configuration for which our neural net perfectly fits the data, however, it is far from obvious why the gradient descent finds this configuration.
\end{enumerate}

In the current essay we will talk about the optimization mystery.

\begin{figure}
    \centering
    \includegraphics[width=0.45\textwidth]{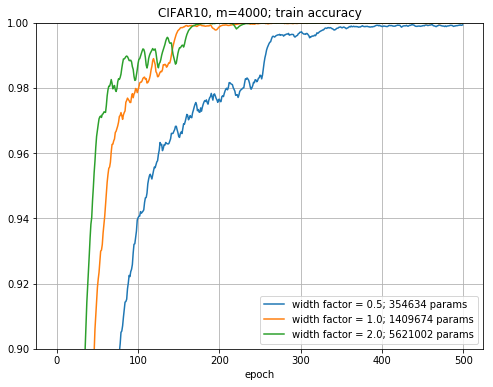}
    \includegraphics[width=0.45\textwidth]{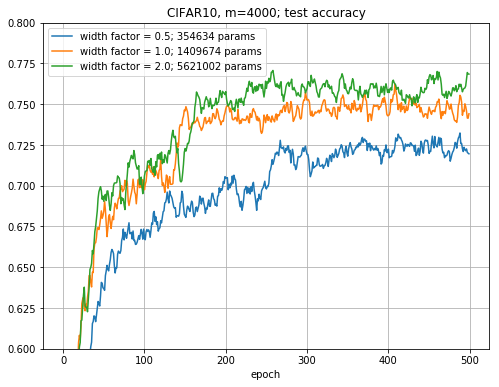}
    \caption{Results of training Conv-small of~\cite{Miyato_et_al_2017} on subset of 4000 samples of CIFAR10. Initial numbers of filters in convolutional layers were multiplied by width factor.}
    \label{fig:results_cifar}
\end{figure}

In order to illustrate the above-mentioned phenomena, we trained a modern convolutional architecture, Conv-small from~\cite{Miyato_et_al_2017}, on a subset of 4000 examples of CIFAR-10 dataset. We also multiplied the number of filters in each convolutional layer by some width factor, in order to illustrate how the number of parameters affects optimization and generalization.

The results are shown in Figure~\ref{fig:results_cifar}. One can see that the network indeed generalizes well, despite of severe overparameterization. Moreover, the test accuracy does not decrease as network grows larger. From the optimization side, we see that SGD indeed finds a global minimum in terms of train error (note that there are no global minima in terms of cross-entropy in any finite ball in weight-space). Moreover, as network grows larger, optimization becomes faster (in terms of number of iterations).

\section{Optimization mystery}

\subsection{Loss landscape}

Let us first deal with one of these observations: gradient descent finds a global minimum for random initialization. It is known for randomly initialized gradient descent that it is able to find a local minimum almost surely (Theorem 4.8 of~\cite{Lee_et_al_2016}). Hence it is tempting to propose the following hypothesis:
\begin{equation}
    \text{All local minima of loss landscape of neural nets are global.}
    \label{loss_landscape_hypothesis}
\end{equation}

We begin with the simplest optimization problem --- $l_2$-regression:
\begin{equation}
    \LL_{shallow}(W) = 
    \frac12 \EE_{x,y \sim \DD} \|y - Wx\|_2^2 \to \min_W.
    \label{l2_regression}
\end{equation}
One can easily show that all local minima of this problem are indeed global, and in the case of non-degenerate data covariance matrix $\Sigma_{xx} = \EE_x xx^T$, this minimum is unique.

There are two possible ways to depart from this trivial setting: acquiring depth, and adding a non-linearity. 

\subsubsection{Deep linear nets}

Let us consider a deep linear net first. The corresponding $l_2$-regression problem is given as follows:
\begin{equation}
    \LL_{deep}(W_{1:H}) = 
    \frac12 \EE_{x,y \sim \DD} \|y - W_H \ldots W_1 x\|_2^2 \to \min_{W_{1:H}}.
    \label{deep_problem}
\end{equation}
This setting looks odd, since in terms of a class of realizable functions, a deep linear net is equivalent to a shallow one with rank restriction. Indeed, let $W_k \in \RR^{d_k \times d_{k-1}}$, and consider the following shallow problem:
\begin{equation}
    \LL_{shallow}(R) = 
    \frac12 \EE_{x,y \sim \DD} \|y - Rx\|_2^2 \to \min_R, \quad \st \; \rk R \leq d_p,
    \label{shallow_problem}
\end{equation}
where $p \in \Argmin_k d_k$ --- index of the "bottleneck" layer.
Then, we have a many-to-one correspondence:
$$
    W_{1:H} \to R = W_H \ldots W_1,
$$
and all minima of a deep problem are of the same value as ones of a shallow problem:
$$
    \min_R \LL_{shallow} = \min_{W_{1:H}} \LL_{deep}.
$$
Note, however, that both problems are (generally) non-convex: shallow problem is non-convex, because of non-convexity of its optimization domain (for non-trivial rank-constraint), and deep problem is non-convex due to weight-space symmetries.
Indeed, if we multiply one of the weight matrices by a non-zero factor, and divide another weight matrix by the same factor simultaneously, the corresponding function will not change.

The above-proposed hypothesis~\ref{loss_landscape_hypothesis} is false for a general non-convex problem, however, it was proven in ~\cite{Lu_Kawaguchi_2017} (see Theorem 2.2 there) that it is true for a rank-constrained shallow problem\footnote{More precisely, Theorem 2.2 of~\cite{Lu_Kawaguchi_2017} assumes a finite dataset. It is not obvious how to generalize it to a general data distribution.}.
Moreover,~\cite{Lu_Kawaguchi_2017} prove that the same is true for a deep problem too.

Consider a finite dataset $X \in \RR^{d_0 \times m}$ of size $m$. Assume $X$ to be of full rank (this corresponds to non-degeneracy of the corresponding covariance matrix). At the high level, the proof is based on three theorems:
\begin{enumerate}
    \item Theorem 2.1: 
    If $W_{1:H}$ is a local minimum of $\LL_{deep}(W_{1:H})$, then $R = W_H \ldots W_1$ is a local minimum of $\LL_{shallow}(R)$.
    \item Theorem 2.2:
    Every local minimum of $\LL_{shallow}(R)$ is global.
    \item Theorem 2.3:
    Every local minimum of $\LL_{deep}(W_{1:H})$ is global.
\end{enumerate}

Note that Theorem 2.3 is a simple corollary of two previous theorems.
Indeed, let $W_{1:H}$ be a local minimum of $\LL_{deep}$, then due to Theorem 2.1, $R = W_H \ldots W_1$ is a local minimum of $\LL_{shallow}$. Since all local minima of $\LL_{shallow}$ are of the same value (Theorem 2.2), all local minima of $\LL_{deep}$ are of the same value too. Hence they are all global.

Proof of Theorem 2.1 starts with the following observation. Let $\bar W_{1:H}$ be a local minimum of $\LL_{deep}$. Let $W_{1:H}$ be a small perturbation of $\bar W_{1:H}$. In order to prove that $\bar R = \bar W_H \ldots \bar W_1$ is a local minimum of $\LL_{shallow}$, we have to show that for any perturbation $R$ of $\bar R$ holds $\LL_{shallow}(R) \geq \LL_{shallow}(\bar R)$. Our aim then is to prove that for any perturbation $R$ there exists a perturbation $W_{1:H}$ such that $R = W_H \ldots W_1$. Once this is proven, we can conclude that:
$$
    \LL_{shallow}(R) =
    \LL_{deep}(W_{1:H}) \geq
    \LL_{deep}(\bar W_{1:H}) =
    \LL_{shallow}(\bar R).
$$

Perturbation $R$ that corresponds to perturbation $W_{1:H}$ is explicitly constructed for the case $\rk \bar R = d_p$, and $H = 2$. Generalization to $H > 2$ is given by induction (Theorem 3.1). If $\rk \bar R < d_p$, it is proven in Theorem 3.2 and Theorem 3.3 that one can perturb $\bar W_{1:H}$ in such a way that perturbation would have $\rk (W_H \ldots W_1) = d_p$, will be a local minimum of $\LL_{deep}$, and loss will not change. In other words, we can always perturb a local minimum to make it full rank and retain local minimality.

\subsubsection{Non-linear nets}

So, our hypothesis is true for linear nets.
However, a linear net is not a practical case, and we have to move to non-linear nets. A general loss of the simplest non-linear net is given as follows:
\begin{equation}
    \LL(W) = \frac12 \EE_{x,y \sim \DD} \|y - \sigma(W x)\|_2^2,
    \label{simplest_non_linear_problem}
\end{equation}
where $\sigma(\cdot)$ is an element-wise non-linearity.
Unfortunately, it happens that even a simple one-layer non-linear regression has a quite hard-to-analyze loss landscape, even for gaussian data distribution and labels generated from a teacher net of the same architecture (see, e.g.~\cite{Tian_2017}). Hence one can hardly hope to obtain a simple, analyzable loss landscape for a non-linear net with one hidden layer:
\begin{equation}
    \LL(W_{1:2}) = \frac12 \EE_{x,y \sim \DD} \|y - W_2 \sigma(W_1 x)\|_2^2.
    \label{shallow_non_linear_problem}
\end{equation}

\paragraph{Wide shallow sigmoid nets.}

Surprisingly, our hypothesis becomes true for a non-linear net as its hidden layer becomes sufficiently wide. More precisely, consider the following variant of the previous problem:
\begin{equation}
    \LL(W_{1:2}) = \frac12 \|Y - W_2 \sigma(W_1 X)\|_F^2 \to \min_{W_{1:2}},
    \label{shallow_non_linear_problem_finite_data}
\end{equation}
where index $F$ denotes Frobenius norm, $X \in \RR^{d_0 \times m}$ is a finite dataset of size $m$, $W_k \in \RR^{d_k \times d_{k-1}}$. Note that here $d_1$ denotes the width of a hidden layer.
\cite{Yu_Chen_1995} proved that as long as $d_1 \geq m$ and columns of $X$ (data points) are distinct, for $\sigma(z) = (1 + \exp(-z))^{-1}$ all local minima are global\footnote{Actually, \cite{Yu_Chen_1995} proved their theorem for the case $d_1 = m$, but it is not hard to generalize it to $d_1 \geq m$.}.

The proof is based on the following observation. Let $W_{1:2}$ be a local minimum of~\ref{shallow_non_linear_problem_finite_data}. Let us fix $W_1$, and consider the following surrogate loss:
$$
    \LL_{W_1}(W_2) =
    \LL(W_{1:2}).
$$
Note that $\LL_{W_1}$ corresponds to the loss of a linear regression~\ref{l2_regression} with modified dataset $F_1 = \sigma(W_1 X) \in \RR^{d_1 \times m}$.
Since the problem
$\LL_{W_1}(W_2) \to \min_{W_2}$
is convex, all of its critical points are minima of the same value.
Obviously, if $F_1$ has rank $m$ (this necessitates the width $d_1$ of a hidden layer to be greater than the number of datapoints $m$), we can perfectly fit the modified dataset, and all local minima of $\LL_{W_1}$ have zero value. In this case, $\LL(W_{1:2}) = \LL_{W_1}(W_2) = 0$, and $W_{1:2}$ is indeed a global minimum of $\LL$.

Unfortunately, the condition $d_1 \geq m$ is not sufficient for $F_1$ to always be of rank $m$. However, it is proven in the paper that as long as all columns of $X$ are distinct, the set of such $W_1$ for which $\rk F_1 < m$, has Lebesgue measure of zero\footnote{\cite{Yu_Chen_1995} gave an incorrect proof of this statement. One can find a correct proof of a more general statement in Lemma 4.4 of~\cite{Nguyen_Hein_2017}}. The proof sufficiently utilizes the fact that sigmoid is an analytic function; the proven statement is false for ReLU.

The proof of the main theorem proceeds as follows. We have already proven the theorem for $\rk F_1 = m$. However, if $\rk F_1 < m$, since the value of the measure for the corresponding $W_1$ is zero, we can find a small enough perturbation $\hat W_1$, so that $\rk \hat F_1 = m$. Suppose $\LL(W_{1:2}) > 0$. Then, for a small enough perturbation $\hat W_1$, $\LL(\hat W_1, W_2) > 0$. Launch a gradient flow on $\LL_{\hat W_1}$. It will find a point $\tilde W_2$ with zero loss, since $\rk \hat F_1 = m$. Hence a gradient flow leaves some fixed vicinity of $W_{1:2}$ (vicinity of positive loss), no matter how small the perturbation $\|\hat W_1 - W_1\|$ is. This means that the point $W_{1:2}$ is unstable in Lyapunov sense, hence it cannot be a local minimum. Hence all local minima of~\ref{shallow_non_linear_problem_finite_data} have value zero, regardless of the rank of $F_1 = \sigma(W_1 X)$; hence all of them are global.

\paragraph{Deep and wide sigmoid nets.}

We see that our hypothesis~\ref{loss_landscape_hypothesis} is true for wide-enough shallow nets. What about deep nets?
Consider a problem, similar to~\ref{shallow_non_linear_problem_finite_data}:
\begin{equation}
    \LL(W_{1:H}) = 
    \|Y - W_H \sigma(W_{H-1} \ldots \sigma(W_1 X) \ldots )\|_F^2 \to \min_{W_{1:H}},
    \label{deep_non_linear_problem_finite_data}
\end{equation}
where as before, $X \in \RR^{d_0 \times m}$ denotes a finite dataset of size $m$, $W_k \in \RR^{d_k \times d_{k-1}}$, and $\sigma(\cdot)$ is an element-wise sigmoid non-linearity.
In the previous theorem the main assumption was that the hidden layer was wide enough. Here we assume a similar restriction:
$$
    \exists k: \; d_k \geq m, \; \text{and all columns of $X$ are distinct.}
$$
These two assumptions are enough to prove that the set of $W_{1:k}$, such that $F_k = \sigma(W_k \ldots \sigma(W_1 X) \ldots ) \in \RR^{d_k \times m}$ has rank $<m$, has Lebesgue measure zero (Lemma 4.4 of~\cite{Nguyen_Hein_2017}).

Let us try directly applying the logic of the previous theorem.
For a shallow net $\rk F_1 = m$ implied $\LL(W_{1:2}) = 0$ at a local minimum. This was due to the fact that the problem $\LL_{W_1}(W_2) = \LL(W_{1:2}) \to \min_{W_2}$ was convex (since it was a linear regression). Unfortunately, given $\rk F_k = m$, the problem $\LL_{W_{1:k}}(W_{k+1:H}) = \LL(W_{1:H}) \to \min_{W_{k+1:H}}$ is not generally convex (since for $H > k+1$ it is not a linear regression any more). Hence it is not true that all local minima have zero loss, even given $\rk F_k = m$.

However, if we additionally assume that our local minimum is non-degenerate in the following sense:
$$
    \forall l \in \{k+2,\ldots,H\} \quad \rk W_l = d_l,
$$
then $\rk F_k = m$ at a local minimum will imply that the loss in this local minimum is zero (Lemma 3.5 of~\cite{Nguyen_Hein_2017}). Note that for a shallow net $k=1$, $H=2$, and this assumption is trivially true. Note also that this assumption implies $d_l \leq d_{l-1} \; \forall l \geq k+2$. In other words, the width of a net should not increase after the $k$-th layer.

The condition $\rk F_k = m$ does not generally hold even given $d_k \geq m$. Similar to the theorem concerning shallow nets, we have to deal with the case when $\rk F_k < m$ at our local minimum. Previously, we have used the convexity of the problem $\LL_{W_1}(W_2) \to \min_{W_2}$, which implied that continuous-time gradient descent finds a global minimum of this problem. Now we have the problem $\LL_{W_{1:k}}(W_{k+1:H}) = \LL(W_{1:H}) \to \min_{W_{k+1:H}}$, which is not generally convex, and we do not have a guarantee to find a global minimum of it with continuous-time gradient descent any more.

In their Theorem 4.6,~\cite{Nguyen_Hein_2017} use the following technique. They assume the Hessian $\nabla^2_{W_{k+1:H}} \LL(W_{1:H})$ to be non-degenerate at a local minimum $W_{1:H}$ we consider. This allows them to use the implicit function theorem to deduce that if $W_{1:H}$ is a critical point, then for any perturbation $\hat W_{1:k}$ one can find a perturbation $\hat W_{k+1:H}$, such that $\hat W_{1:H}$ is also a critical point.
Take a perturbation $\hat W_{1:k}$ such that $\hat F_k$ has rank $m$ (such perturbation exists due to Lemma 4.4, mentioned previously). The corresponding perturbed matrices $\hat W_l$ for $l \geq k+2$ will have ranks $d_l$ as long as perturbations are small enough. Hence at the perturbed point $\hat W_{1:H}$ we find ourselves at a simple, previously-discussed case, and conclude that $\LL(\hat W_{1:H}) = 0$. However, since the loss function is continuous wrt weights, we further conclude that $\LL(W_{1:H}) = 0$, and the considered minimum is global.

To sum up, the main theorem (Corollary 3.9 of~\cite{Nguyen_Hein_2017}) states the following. Let $W_{1:H}$ be a local minimum of a problem~\ref{deep_non_linear_problem_finite_data}. Assume the following conditions hold:
\begin{enumerate}
    \item Columns of $X$ (data points) are distinct;
    \item One of the layers is wide enough: $\exists k: \; d_k \geq m$;
    \item Our minimum is non-degenerate: $\forall l \in \{k+2,\ldots,H\} \quad \rk W_l = d_l$;
    \item Loss function has non-degenerate Hessian at our minimum: $\det \nabla^2_{W_{k+1:H}} \LL(W_{1:H}) \neq 0$.
\end{enumerate}
Then, $\LL(W_{1:H}) = 0$, and $W_{1:H}$ is a global minimum of~\ref{deep_non_linear_problem_finite_data}.

Unfortunately, as we see, this theorem does not state that all local minima of~\ref{deep_non_linear_problem_finite_data} are global. It states only that some minima, which are non-degenerate in some sense, are global. We hypothesize that the third condition could be relaxed to simply $d_l \leq d_{l-1} \; \forall l \in \{k+2,\ldots,H\}$, but we also suspect that degenerate minima in terms of the Hessian can exist, and our initial hypothesis~\ref{loss_landscape_hypothesis} is generally false.

\paragraph{Remark on non-smooth non-linearities.}

The above theorem could be generalized to all commonly-used \emph{smooth} non-linearities, e.g. tanh, softplus, however, it is unlikely for it to be generalized to non-smooth ones: ReLU, or LeakyReLU. Indeed, when some units of a ReLU network are deactivated, we effectively obtain a smaller network. We do not have guarantees for not-wide-enough nets. Hence, in the worst case, a smaller network could have local minima. These local minima could become (degenerate) saddles in the initial network, or again (degenerate) local minima. It is not obvious, whether the second alternative is possible or not. The same reasoning holds for the theorem of~\cite{Yu_Chen_1995}.

\paragraph{Remark on criticality with respect to $W_{1:k}$.}

One can notice that in the proof of the theorem of~\cite{Nguyen_Hein_2017} we have never used the fact that the local minimum $W_{1:H}$ is a critical point of loss function $\LL$ wrt the first $k$ weight matrices, i.e. that $\nabla_{W_{1:k}} \LL(W_{1:H}) = 0$. It means that given the four above-stated conditions, we only need to assume $W_{k+1:H}$ to be a local minimum of $\LL_{W_{1:k}}(W_{k+1:H}) = \LL(W_{1:H})$. Hence, according to the theorem, $W_{1:k}$ becomes a minimum of $\LL_{W_{k+1:H}}(W_{1:k})$ automatically. It is worth thinking of how the condition $\nabla_{W_{1:k}} \LL(W_{1:H}) = 0$ could be used to relax some of the four above-stated assumptions. The same reasoning holds for theorem of~\cite{Yu_Chen_1995}.

\subsection{Gradient descent dynamics}

As we have seen in the previous section, despite the fact, that gradient descent is guaranteed to converge to a local minimum, it is not obvious, whether all local minima are global or not, especially for ReLU networks, where theorem of~\cite{Nguyen_Hein_2017} does not hold. However, as we can see in Figure~\ref{fig:results_cifar}, (stochastic) gradient descent indeed converges to a global minimum on the train set.

A prominent result in this direction was obtained by~\cite{Du_et_al_2018}. Consider a ReLU net with one hidden layer and a single output:
\begin{equation}
    f(W, a, x) = 
    \frac{1}{\sqrt{m}} \sum_{r=1}^m a_r [w_r^T x]_+,
    \label{shallow_relu_net}
\end{equation}
where $x \in \RR^d$ is an input, $w_r \in \RR^d$ are weights of the hidden layer, $a_r \in \RR$ are weights of the output layer, and $[z]_+ = \max(0,z)$ denotes an element-wise ReLU. Note that the width of the hidden layer is denoted by $m$ here. We consider $l_2$ regression on a finite dataset of size $n$. This leads to the following loss function:
\begin{equation}
    L(W) =
    \frac12 \sum_{i=1}^n (f(W, a, x_i) - y_i)^2.
    \label{shallow_relu_net_l2_loss}
\end{equation}
Note that the corresponding optimization problem $L(W) \to \min_W$ is very similar to~\ref{shallow_non_linear_problem_finite_data}. Our new problem differs significantly from~\ref{shallow_non_linear_problem_finite_data} in the following aspects:
\begin{enumerate}
    \item ReLU instead of sigmoid,
    \item optimization is performed wrt to $W$ only.
\end{enumerate}
Note also the change of notation (in order to be compatible with the paper).

The result of~\cite{Du_et_al_2018} easily transfers to sigmoid, but it is unlikely that the result of~\cite{Yu_Chen_1995} could be generalized to ReLU. Hence the first point is by no means a drawback, it is an advantage.

The last point is even more crucial. The proof of theorem of~\cite{Yu_Chen_1995} critically relied on optimization wrt $W_2$ (which is denoted as $a$ here). The proof of \cite{Du_et_al_2018} does not rely on it. The main result (which we have not stated yet), could be obtained with or without optimization wrt $a$.

Before stating the main result of~\cite{Du_et_al_2018}, we need several assumptions and definitions.
Assume that the weights are initialized as follows:
$$
    w_r(0) \sim \NN(0, I), \; a_r \sim U(\{-1, 1\}) 
    \quad \forall r = 1, \ldots, m.
$$
Assume also that datapoints lie on a sphere and y's are bounded:
$$
    \|x_i\|_2 = 1, \; |y_i| < C \quad \forall i = 1,\ldots,n.
$$
We consider dynamics of a continuous-time gradient descent:
$$
    \frac{dw_r(t)}{dt} =
    -\frac{\partial L(W(t))}{\partial w_r}.
$$
In the proof we are going to reason about the dynamics of network individual predictions:
$$
    u_i(t) =
    f(W(t), a, x_i).
$$

We now proceed to the main result.
Theorem 3.2 of~\cite{Du_et_al_2018} states the following.
Let $\delta \in (0, 1)$ and $m = \Omega\left(\frac{n^6}{\lambda_0^4 \delta^3}\right)$; then w.p. $\geq 1-\delta$ over initialization we have:
\begin{equation}
    \|u(t) - y\|_2^2 \leq
    e^{-\lambda_0 t} \|u(0) - y\|_2^2 \quad 
    \forall t \geq 0.
    \label{du_th32}
\end{equation}
Here $\lambda_0$ is a data-dependent constant, which we will define soon. Theorem 3.1 of~\cite{Du_et_al_2018} states that as long as there are no data points which are parallel to each other, this constant is strictly positive.

Let us take a closer look at this result. It states basically that as long as the hidden layer is wide enough, neural network predictions on data-points converge to ground-truth answers exponentially fast with high probability. A trivial consequence is that continuous-time gradient descent indeed converges to global minimum with high probability.

Note that the continuous-time gradient descent is assumed only for convenience. A similar result (with proper learning rate) could be obtained for discrete-time gradient descent as well (Theorem 4.1 of~\cite{Du_et_al_2018}).

\paragraph{Proof sketch.}

The proof starts with the following reasoning. We want to know how network predictions change with time:
\begin{equation*}
    \begin{split}
        \frac{du_i(t)}{dt} &=
        \sum_{r=1}^m \frac{\partial u_i(t)}{\partial w_r} \frac{dw_r}{dt} =
        -\sum_{r=1}^m \frac{\partial u_i(t)}{\partial w_r} \frac{\partial L(W(t))} {\partial w_r} =\\&=
        \sum_{j=1}^n (y_j - u_j(t)) \sum_{r=1}^m \left\langle \frac{\partial u_i(t)}{\partial w_r}, \frac{\partial u_j(t)}{\partial w_r} \right\rangle =\\&=
        \sum_{j=1}^n (y_j - u_j(t)) \frac{1}{m} x_i^T x_j \sum_{r=1}^m a_r^2 [w_r^T x_i \geq 0, w_r^T x_j \geq 0],
    \end{split}
\end{equation*}
where square brackets denote indicators. Note that $a_r^2 = 1$ at initialization, and does not change during training process. Hence we shall omit it from now on. If we define
\begin{equation}
    H_{ij}(t) =
    \frac{1}{m} x_i^T x_j \sum_{r=1}^m [w_r^T x_i \geq 0, w_r^T x_j \geq 0],
    \label{H_definition}
\end{equation}
then we can rewrite the previous equation as:
$$
    \frac{du(t)}{dt} =
    H(t) (y-u(t)).
$$
This differential equation governs the evolution of network predictions on train set. One can easily show that $H(t)$ is a Gram matrix, hence it is positive semi-definite. Our main goal is to show that its minimal eigenvalue remains bounded below with $\lambda_0/2$ throughout the whole process of optimization:
\begin{equation}
    \text{Our goal is to prove $\lambda_{min}(H(t)) \geq \frac{\lambda_0}{2} \quad \forall t \geq 0$.}
    \label{du_proof_goal}
\end{equation}
If this holds, we easily obtain the desired result:
\begin{equation*}
    \begin{split}
        \frac{d}{dt} \|y - u(t)\|_2^2 &=
        -2 (y - u(t))^T \frac{du(t)}{dt} =\\&=
        -2 (y - u(t))^T H(t) (y - u(t)) \leq
        -\lambda_0 \|y- u(t)\|_2^2.
    \end{split}
\end{equation*}
After solving the differential inequality we get:
$$
    \|y - u(t)\|_2^2 \leq 
    e^{-\lambda_0 t} \|y - u(0)\|_2^2,
$$
which is~\ref{du_th32}.
Before giving the sketch of the proof, we have to define $\lambda_0$.
Let $H^\infty$ be the expectation of Gram matrix at initialization $H(0)$:
$$
    H_{ij}^\infty =
    \EE_{w_1 \sim \NN(0,I)} \ldots \EE_{w_r \sim \NN(0,I)} H_{ij}(0) =
    \EE_{w \sim \NN(0, I)} (x_i^T x_j [w^T x_i \geq 0, w^T x_j \geq 0]).
$$
We define $\lambda_0$ as its minimal eigenvalue, which is non-negative, since $H^\infty$ is a Gram matrix. Theorem 3.1 of~\cite{Du_et_al_2018} states that it is positive as long as data-points are not parallel to each other.

The proof of~\ref{du_proof_goal} is divided into two parts. 
In the first part we prove that as long as $m$ is large enough, $H(0)$ is close to its expectation, $H^\infty$, in terms of $l_2$ norm. As a consequence, the spectrum of $H(0)$ does not differ much from the spectrum of $H^\infty$, and $\lambda_{min}(H(0)) \geq \frac34 \lambda_0$.
In the second part we prove that as long as $m$ is large enough, $H(t)$ does not change much in terms of $l_2$ norm throughout the optimization process. As a consequence, its spectrum also does not change much, and $\lambda_{min}(H(t)) \geq \frac12 \lambda_0 \; \forall t \geq 0$.

The first part is covered by Lemma 3.1 of~\cite{Du_et_al_2018}. Essentially, it uses a concentration bound (Hoeffding inequality) to bound $|H_{ij}(0) - H_{ij}^\infty|$, and then to bound $\|H(0) - H^\infty\|_2$.

The second part is a bit more involved. 
Lemma 3.2 of~\cite{Du_et_al_2018} states that as long as weights are close enough to initialization, Gram matrix does not change much. This is due to the fact that if we perturb the initial weights, few of the indicators in the definition~\ref{H_definition} change.
Lemma 3.3 states that as long as $\lambda_{min}(H(s)) \geq \lambda_0/2$ on $[0,t]$, $W(t)$ is sufficiently close to $W(0)$.
Lemmas 3.2 and 3.3 together imply that $\lambda_{min}(H(s)) \geq \lambda_0/2$ on $[0,t]$ is equivalent to the fact that $W(s)$ is close to $W(0)$ on $[0,t]$.
Since for $t=0$ both statements should necessarily be true, Lemma 3.4 states that they are true for all $t \geq 0$ using continuity argument.

\paragraph{Remark on convergence rate.}

The theorem~\ref{du_th32} states that neural network predictions on train dataset converge to train labels exponentially fast, which is a very strong result.
However, in optimization theory convergence rate is usually measured in terms of how fast the algorithm converges to a stationary point.
A (first-order) stationary point is a critical point of loss function:
$$
    \nabla_W L(W) = 0.
$$
A second-order stationary point is a stationary point with positive semi-definite hessian of loss function:
$$
    \nabla_W L(W) = 0; \quad \nabla^2_W L(W) \geq 0.
$$

It is well known in optimization theory that given a general gradient-Lipschitz loss function $L: \, \RR^d \to \RR$ gradient descent converges to a stationary point in time independent from the number of parameters $d$.
Let us look at how fast the gradient of loss function decays to zero in our case:
\begin{equation*}
    \begin{split}
        \left\| \frac{\partial L(W(t))}{\partial w_r} \right\|_2 &=
        \frac{1}{\sqrt{m}} \left\| \sum_{j=1}^n (u_j(t) - y_j) a_r x_j [w_r^T(t) x_j \geq 0] \right\|_2 \leq\\&\leq
        \frac{1}{\sqrt{m}} \sum_{j=1}^n |u_j(t) - y_j| =
        \frac{1}{\sqrt{m}} \|u(t) - y\|_1 \leq\\&\leq
        \sqrt{\frac{n}{m}} \|u(t) - y\|_2 \leq
        \sqrt{\frac{n}{m}} e^{-\lambda_0 t / 2} \|u(0) - y\|_2.
    \end{split}
\end{equation*}
We see that as $m$ increases (which is related to the number of parameters), gradient norm decreases, hence the time to reduce the gradient norm to a value below some epsilon also decreases, which was not the case in general setting.

What about second-order stationary points? Theorem 4.8 of~\cite{Lee_et_al_2016} states that randomly initialized gradient descent converges to a second-order stationary point almost surely. However, this theorem tells nothing about the convergence rate. \cite{Du_et_al_2017} constructs an example of loss function $L: \, \RR^d \to \RR$ for which randomly initialized gradient descent requires time exponential in $d$ in order to converge to a second-order stationary point.
It is not hard to check that in our case the Hessian of the loss function is positive semi-definite almost everywhere (except the points where ReLU is not differentiable). Indeed, let us take a look at our Hessian:
\begin{equation*}
    \begin{split}
        \frac{\partial^2 L(W(t))}{\partial w_r \partial w_s} &=
        \frac{1}{m} \sum_{j=1}^n a_r a_s x_j^2 [w_r^T(t) x_j \geq 0, w_s^T(t) x_j \geq 0] =\\&=
        \frac{1}{m} \sum_{j=1}^n \left\langle a_r [w_r^T(t) x_j \geq 0] x_j, a_s [w_s^T(t) x_j \geq 0] x_j \right\rangle.
    \end{split}
\end{equation*}
We see that the Hessian is a sum of Gram matrices, which are positive semi-definite, hence the whole Hessian is positive semi-definite too.
Given that, our method converges to a second-order stationary point in time that decreases with the number of parameters.

\section{Conclusion}

As we have seen, the optimization mystery could be revealed, but either for linear networks (theorem of~\cite{Lu_Kawaguchi_2017}), which are not commonly used, or for unrealistically wide non-linear nets (theorems of \cite{Yu_Chen_1995}, \cite{Nguyen_Hein_2017}, and \cite{Du_et_al_2018}).
We suspect that assuming a cluster structure of the dataset (i.e. classes of MNIST are separated from each other in pixel space) will allow to reduce the required amount of overparameterization sufficiently (this idea was actually elaborated in~\cite{Nguyen_Hein_2017}).
However, it seems that even for unstructured data, the amount of overparameterization required in theorem of ~\cite{Du_et_al_2018} is still too large~\cite{Zhang_et_al_2016}.
We suspect that these bounds could be reduced even in general case using more involved techniques.

\bibliographystyle{apalike}
\bibliography{bibliography}

\end{document}